\documentclass{article}

\usepackage{microtype}
\usepackage{graphicx}
\usepackage{subcaption}
\usepackage{booktabs} % for professional tables
\usepackage{graphicx}
\usepackage{booktabs}
\usepackage{amsmath}
\usepackage{amssymb}
\usepackage{hyperref}
\usepackage{microtype}
\usepackage{xcolor}
\usepackage{multirow}
\usepackage{array}
\usepackage{comment}
\usepackage{float}
\usepackage{subcaption}
\usepackage{colortbl}
\usepackage{hyperref}

\usepackage[preprint]{icml2026}

\usepackage{amsmath}
\usepackage{amssymb}
\usepackage{mathtools}
\usepackage{amsthm}

\usepackage[capitalize,noabbrev]{cleveref}

\theoremstyle{plain}

\theoremstyle{definition}

\theoremstyle{remark}

\usepackage[textsize=tiny]{todonotes}

\icmltitlerunning{Architecture, Not Scale: Circuit Localization in Large Language Models}

\begin{document}

\twocolumn[
  \icmltitle{Architecture, Not Scale: Circuit Localization in Large Language Models}

  \icmlsetsymbol{equal}{*}

  \begin{icmlauthorlist}
    \icmlauthor{Sohan Venkatesh}{mit}
  \end{icmlauthorlist}
    
  \icmlaffiliation{mit}{Manipal Institute of Technology Bengaluru}
    
  \icmlcorrespondingauthor{Sohan Venkatesh}{soh.venkatesh@gmail.com}

  \vskip 0.3in
]



\printAffiliationsAndNotice{}  % no special notice (required even if empty)

\begin{abstract}
Mechanistic interpretability assumes that circuit analysis becomes
harder as models scale. We challenge this assumption by showing that the
attention architecture matters more than parameter count. Studying three
circuit types across Pythia and Qwen2.5, we find that grouped query
attention produces circuits that are far more concentrated and
mechanistically stable than standard multi-head attention at comparable
scales. The same concentration pattern holds across indirect object
identification, induction heads and factual recall. Within a single architecture 
family (Qwen2.5), factual recall circuits undergo a discrete phase transition above a critical scale, 
collapsing to a single bottleneck rather than degrading gradually. These findings 
suggest that some architectural choices make large models more tractable to study and that
interpretability difficulty is not a fixed consequence of model size.
\end{abstract}

\section{Introduction}
\label{sec:intro}
Mechanistic interpretability aims to reverse-engineer neural networks
into understandable components such as circuits, features and
representations that explain specific model behaviors
\citep{olah2020zoom, elhage2021mathematical}. The core premise is that
models encode computations in identifiable structures that can be
located, ablated and understood. This has proven productive on small
models
\citep{olsson2022context, wang2022interpretability, meng2022locating} but
a practical concern remains: does mechanistic interpretability stay feasible as models
scale to billions of parameters?
 
The standard assumption is that it does not. Larger models are expected
to develop more redundant representations, distribute computation across
more components and resist the surgical ablations that make small-model
circuits legible \citep{lindsey2025biology, elhage2022toy}. This
belief shapes how the field allocates effort, prioritising small
tractable models and developing automated tools designed to cope with
future scale \citep{conmy2023towards}. The assumption has rarely been tested directly
under controlled conditions.  Prior work studying interpretability at scale
has not controlled for architecture as an independent variable \citep{lieberum2023does}.
 
We isolate one key variable, the attention mechanism. We compare Pythia
\citep{biderman2023pythia}, which uses standard Multi-Head Attention
(MHA) throughout, against Qwen2.5 \citep{qwen2_5_technical_report},
which uses Grouped Query Attention (GQA) throughout. We test three circuit 
types (indirect object identification, induction heads and factual recall) 
across six model sizes from 160M to 7B parameters using TransformerLens \citep{nanda2022transformerlens}.
 
Architecture predicts circuit geometry more reliably than scale. GQA
models produce circuits that concentrate into one or two heads across
all three tasks. MHA models produce circuits that spread across tens to
hundreds of heads. The difference follows from the structural
constraints GQA imposes on value-space computation. Ablating one KV head
disrupts all query heads sharing it, creating a bottleneck with no
analogue in MHA.
 
GQA circuits are also mechanistically stable. The same head dominates
regardless of task difficulty or input distribution. MHA circuits shift
substantially between easy and hard input conditions, with the top
contributing heads changing across regimes. This stability asymmetry
matters for safety monitoring. Consistent circuit behavior across inputs
is a prerequisite for reliable oversight \citep{ganguli2022red}.

\section{Related Work}
 
\paragraph{Induction heads.} \citet{olsson2022context} identified
induction heads as a key mechanism for in-context learning across
transformer architectures. These heads implement a pattern-completion
operation: given a repeated sequence $[A][B]\ldots[A]$, they attend back
to the first $A$ and copy the subsequent $B$. We measure how induction
circuit geometry changes with scale and architecture.
 
\paragraph{Indirect object identification.}
\citet{wang2022interpretability} identified the IOI circuit in GPT-2
small using activation patching, characterising name mover heads, backup
name mover heads and inhibition heads. IOI and induction heads are
distinct circuit types. IOI requires semantic name tracking across
sentence structure and routing of the object name to the prediction
position. Induction heads implement a mechanical copy operation that
does not require semantic understanding \citep{mcdougall2024copy}. We
use IOI as our primary evaluation task and extend it across two
architecture families.
 
\paragraph{Factual recall.} \citet{meng2022locating} located factual
associations in mid-to-late MLP layers using causal tracing.
\citet{geva2023dissecting} characterised the role of attention heads in
routing subject information to the final token position. Earlier work by
\citet{geva2021transformer} established that feed-forward layers
function as key-value memories.
 
\paragraph{Scaling and interpretability.} \citet{conmy2023towards}
proposed Automatic Circuit DisCovery (ACDC), which uses iterative 
activation patching to identify the minimal computational subgraph 
implementing a target behaviour. \citet{lieberum2023does}
tested whether circuit analysis scales to Chinchilla-scale models and
found mixed evidence. Standard techniques transferred to the 70B model 
but semantic understanding of the identified components remained partial.
\citet{lindsey2025biology} find that circuits in
larger language models are denser and harder to isolate. We ask whether
architectural choices can produce large models tractable by existing
methods.
 
\paragraph{Features and representations.} \citet{elhage2022toy} showed
that networks store more features than dimensions through superposition.
\citet{templeton2024scaling} showed sparse autoencoders decompose these
into interpretable features at scale. \citet{marks2023geometry} found
emergent linear structure in truth representations.
\citet{hernandez2023linearity} showed relation decoding is linear in
transformer representations.

\section{Background}
 
\paragraph{Induction heads.} An induction head attends from a repeated
token $[A]$ back to its previous occurrence and copies the subsequent
token as its prediction. This enables in-context learning: the model completes novel
patterns seen earlier in context \citep{olsson2022context}. Induction heads typically operate as a
two-head system. A previous-token head shifts attention back one
position and an induction head uses this signal to attend to the token
that followed the prior occurrence.
 
\paragraph{Indirect Object Identification.} The IOI task requires
identifying the recipient in sentences such as ``After Mary and John
went to the store, John gave a mango to \_\_\_'', where the correct
answer is Mary. This differs fundamentally from induction: the model
must track two names, determine their semantic roles and route the
correct name to the prediction position. Wang et
al.~\citeyearpar{wang2022interpretability} decomposed this circuit into
name mover heads, backup name mover heads and inhibition heads. We
measure circuit geometry using logit difference $\text{logit}(\text{IO})
- \text{logit}(\text{S})$ at the final token, where IO is the recipient
and S is the giver.
 
\paragraph{Factual recall.} Factual recall refers to completing
subject-relation-object associations stored in model weights, for
example completing ``The capital of France is'' with ``Paris''. The
circuit involves subject token processing and attention heads that route
subject information to the final token position
\citep{geva2023dissecting}. Prior work using causal tracing
\citep{meng2022locating} located factual associations in mid-to-late
MLP layers of MHA models.
 
\paragraph{Grouped Query Attention.} Standard MHA
\citep{cordonnier2020multi} gives each attention head independent query,
key and value matrices. With $h$ heads, head $i$ computes:
\begin{equation}
  \text{Attn}_i = \text{softmax}\!\left(\frac{Q_i K_i^\top}{\sqrt{d_{\text{head}}}}\right) V_i
\end{equation}

GQA shares key and value matrices across groups of query heads, with
$n_{\text{kv}} < h$ KV heads \citep{ainslie2023gqa}:
\begin{equation}
  \text{Attn}_i = \text{softmax}\!\left(
    \frac{Q_i K_{\lfloor i/r \rfloor}^\top}{\sqrt{d_{\text{head}}}}
  \right) V_{\lfloor i/r \rfloor}, \quad r = h / n_{\text{kv}}
\end{equation}
This reduces KV cache size by $r$ and concentrates value-space computation into 
$n_{\text{kv}}$ shared subspaces. A single KV head mediates the output of all 
$r$ query heads assigned to it.

\section{Methodology}
 
\subsection{Models}
 
We study two architecture families. Pythia comprises Pythia-160M (12 layers, 12 heads), 
Pythia-1.4B (24 layers, 16 heads) and Pythia-6.9B (32 layers, 32 heads). Qwen2.5 
comprises Qwen2.5-0.5B (24 layers, 14 Q-heads, 2 KV-heads), Qwen2.5-1.5B (28 layers, 
12 Q-heads, 2 KV-heads) and Qwen2.5-7B (28 layers, 28 Q-heads, 4 KV-heads).
 
\subsection{Tasks}
We study three circuit types that are well-established in the
mechanistic interpretability literature and each probe a distinct kind
of computation.
 
Indirect Object Identification (IOI) is our primary task. IOI has a
known circuit structure from prior work
\citep{wang2022interpretability}, making it the strongest test of
whether architecture affects circuit geometry on a well-characterised
semantic task. We use the fahamu/ioi dataset
\citep{fahamu2023ioi}, containing 26M IOI sentences. We sample 500
sentences per model and filter to ensure both names tokenise to exactly
one token, preventing multi-token ambiguity in the logit difference
metric. We score each (layer, head) pair using 20 sentences by
measuring the logit difference drop when that head is ablated. Ablation
curves run over the full 500-sentence set.
 
Induction heads serve as a robustness check on a task with no semantic
content. If the same concentration pattern appears on synthetic
repeated-token sequences as on IOI, it is unlikely to be specific to
name-tracking. We construct 200 random repeated-token sequences of the
form $[\text{prefix}][A][B][\text{suffix}][A]$ and measure ICL loss:
the cross-entropy of predicting $[B]$ at the final position. For each
(layer, head) pair we measure mean attention weight at the $AB$ offset
position and run greedy ablation curves.
 
Factual recall uses a curated set of 493 subject-completion facts
spanning ten domains. We build a custom set rather than using existing
benchmarks because TriviaQA and similar datasets contain multi-token
answers and multi-hop chains that complicate single-head ablation
analysis. All facts in our set have single-token answers and known
subject-relation-object structure. Pythia prompts use natural
completion format. Qwen2.5 prompts use a QA format that reliably elicits
factual answers from instruction-aware models. We apply top-3 filtering
for both families. We run two conditions: a per-model condition using
facts each model answers correctly and a shared condition using the
intersection of facts known by all models in a family, which controls
for fact difficulty when comparing across scales.

\subsection{Metrics}
 
We report two main metrics. Top head score is the logit or accuracy drop from 
ablating the single most important head. Higher values indicate a more dominant single 
head. Heads-to-80\% counts the greedy ablations needed to cause 80\% task damage. 
Lower values indicate a more concentrated circuit.

\section{Results}
 
\subsection{GQA Concentrates IOI Circuits into One Head}
 
Table~\ref{tab:ioi} reports IOI results across both families. All six
models solve the task with positive baseline logit difference,
confirming task competence before circuit analysis.

\begin{table*}[t]
\centering
\caption{IOI results across Pythia (MHA) and Qwen2.5 (GQA).
Baseline logit diff measures model confidence in the correct recipient.
Top head score is the logit diff drop from ablating the single most
important head.}
\label{tab:ioi}
\small
\begin{tabular}{lcccc}
\toprule
\textbf{Model} & \textbf{Baseline Logit Diff} & \textbf{Top Head Score} & \textbf{Top Head Layer} & \textbf{Heads-to-80\%} \\
\midrule
Pythia-160M  & 0.290 & 0.108 & L8  & 5 \\
Pythia-1.4B  & 0.408 & 0.231 & L12 & 2 \\
Pythia-6.9B  & 0.546 & 0.250 & L16 & 3 \\
\midrule
Qwen2.5-0.5B & 0.389 & 0.772 & L23 & 1 \\
Qwen2.5-1.5B & 1.368 & 1.860 & L0  & 1 \\
Qwen2.5-7B   & 0.791 & 0.948 & L0  & 1 \\
\bottomrule
\end{tabular}
\end{table*}

\begin{figure*}[t]
\centering
\begin{subfigure}[t]{0.49\linewidth}
  \centering
  \includegraphics[width=\linewidth, trim= 37 37 23 97, clip]{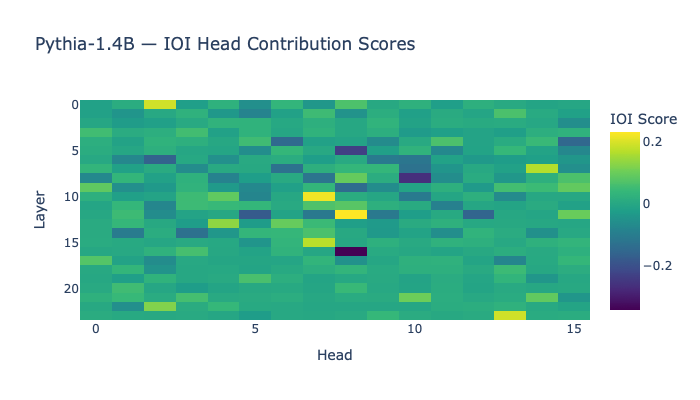}
\end{subfigure}
\hfill
\begin{subfigure}[t]{0.49\linewidth}
  \centering
  \includegraphics[width=\linewidth, trim= 37 37 23 97, clip]{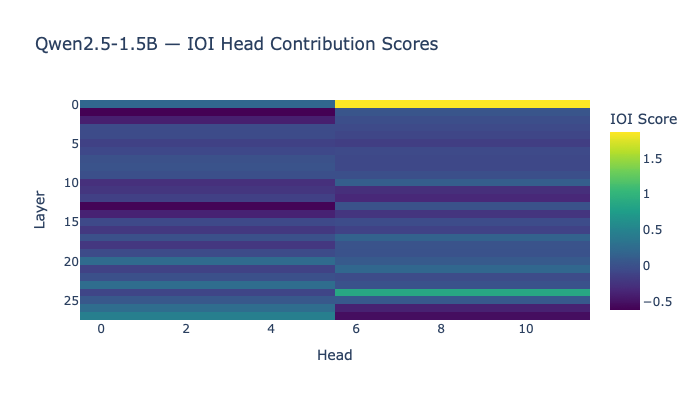}
\end{subfigure}
\caption{IOI head contribution score heatmaps for Pythia-1.4B (left)
and Qwen2.5-1.5B (right) at matched scales. Each cell is the logit diff
drop when that (layer, head) pair is ablated. Pythia-1.4B shows diffuse
contributions spread across many layers and heads. Qwen2.5-1.5B shows a
single bright band at layer 0, a direct consequence of GQA sharing KV heads across all query heads in that layer.}
\label{fig:ioi_heatmaps}
\end{figure*}

\begin{figure*}[t]
\centering
\begin{subfigure}[t]{0.49\linewidth}
  \centering
  \includegraphics[width=\linewidth, trim= 30 37 37 105, clip]{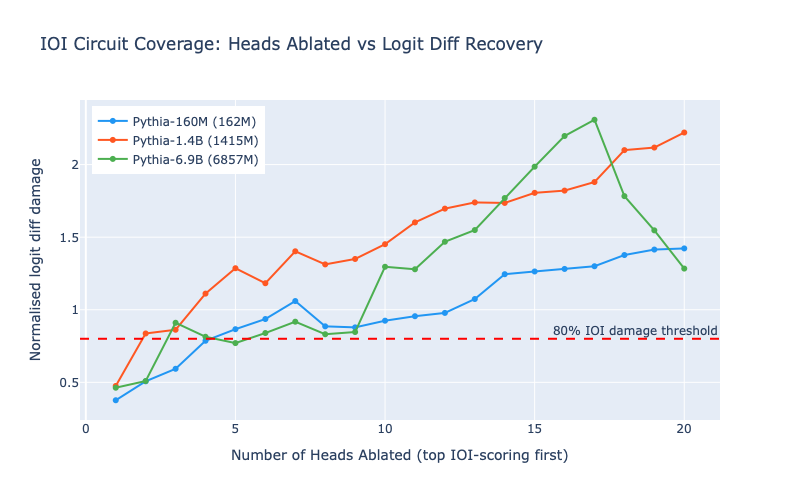}
\end{subfigure}
\hfill
\begin{subfigure}[t]{0.49\linewidth}
  \centering
  \includegraphics[width=\linewidth, trim= 30 37 37 105, clip]{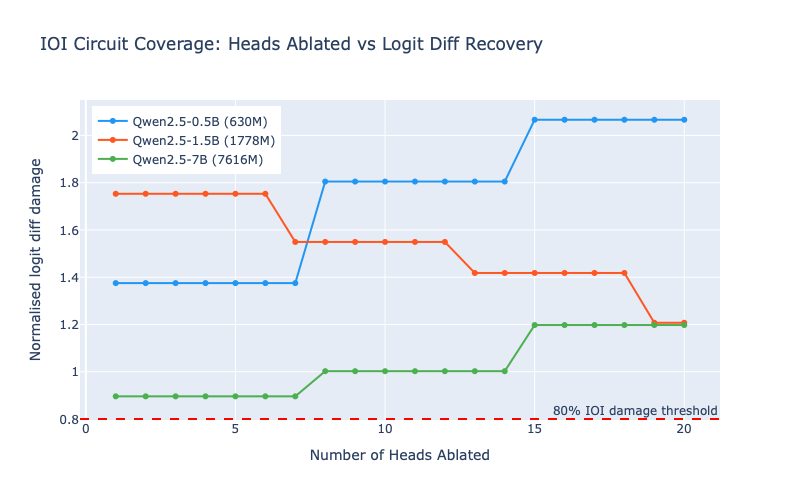}
\end{subfigure}
\caption{IOI ablation curves for Pythia (left) and Qwen2.5 (right).
The y-axis is normalised logit diff damage. The x-axis is heads ablated
in greedy order. Pythia models require multiple ablations before
crossing the 80\% threshold. All three Qwen2.5 models exceed 80\%
damage after the first ablation and remain there which confirms that a single
head carries the circuit.}
\label{fig:ioi_curves}
\end{figure*}

All three Qwen2.5 models require one ablation for 80\% damage while
Pythia requires two to five. Qwen2.5 top head scores are four to eight
times higher than Pythia at comparable scales. Pythia top head layers
shift progressively deeper with scale (L8, L12, L16). Figure~\ref{fig:ioi_heatmaps} shows head contribution score heatmaps at matched scales and Figure~\ref{fig:ioi_curves} shows how logit diff damage accumulates as heads are ablated in greedy order.

Table~\ref{tab:ioi_diagnostic} shows that ablating the single top head alone causes damage comparable to the full greedy sequence, ruling out greedy ordering as the source of the heads-to-80\% result.

\begin{table}[H]
\centering
\caption{Single-head necessity check for Qwen2.5 IOI circuits. Logit Diff (post-ablation) is the logit difference after ablating the top-scoring head alone. Drop (\%) is the percentage reduction from the baseline logit difference.}
\label{tab:ioi_diagnostic}
\small
\begin{tabular}{lcc}
\toprule
\textbf{Model} & \textbf{Logit Diff (post-ablation)} & \textbf{Drop (\%)} \\
\midrule
Qwen2.5-0.5B & $-0.146$ & 137.5 \\
Qwen2.5-1.5B & $-1.030$ & 175.3 \\
Qwen2.5-7B   & $+0.082$ & 89.6  \\
\bottomrule
\end{tabular}
\end{table}
For Qwen2.5-0.5B and 1.5B, ablating the single top head causes
logit difference to flip sign. The model actively predicts the wrong
name after ablation. For Qwen2.5-7B, ablating the top head causes
89.6\% damage. Ablating a randomly chosen mid-layer head as a negative
control causes no damage and in several cases improves logit
difference. This confirms the effect is circuit-specific and not a general
consequence of value zeroing.
 
Qwen2.5-0.5B concentrates at layer 23 while Qwen2.5-1.5B and 7B
concentrate at layer 0. This shift mirrors the phase transition in
factual recall and points to a consistent architectural threshold above
which GQA circuits reorganise to the earliest attention layer.
 
\subsection{Induction Circuit Concentration Depends on Architecture, Not Scale}
 
We run a second experiment on random repeated-token sequences to test whether the IOI concentration pattern holds on a task with no semantic content. We measure ICL loss and score each head by its contribution to in-context prediction. Table~\ref{tab:induction} shows results across both families.

\begin{table}[h]
\centering
\caption{Induction head results on random repeated-token sequences. ICL advantage is random-chance loss minus baseline loss. Pythia-160M has negative ICL advantage and no measurable induction circuit. Ablating up to 20 heads causes no meaningful damage, so its Heads-to-80\% entry exceeds 20.}
\label{tab:induction}
\small
\setlength{\tabcolsep}{2.5pt}
\begin{tabular}{lccc}
\toprule
\textbf{Model} & \textbf{ICL Adv} & \textbf{Top Head Score} & \textbf{Heads-to-80\%} \\
\midrule
Pythia-160M  & $-7.668$ & 0.672 & $>$20 \\
Pythia-1.4B  & $+5.845$ & 0.877 & 22 \\
Pythia-6.9B  & $+5.565$ & 0.766 & 28 \\
\midrule
Qwen2.5-0.5B & $+6.456$ & 0.912 & 2  \\
Qwen2.5-1.5B & $+7.388$ & 0.965 & 6  \\
Qwen2.5-7B   & $+6.278$ & 0.869 & 6  \\
\bottomrule
\end{tabular}
\end{table}
 
\begin{figure*}[t]
\centering
\begin{subfigure}[t]{0.49\linewidth}
  \centering
  \includegraphics[width=\linewidth, trim=30 37 37 105, clip]{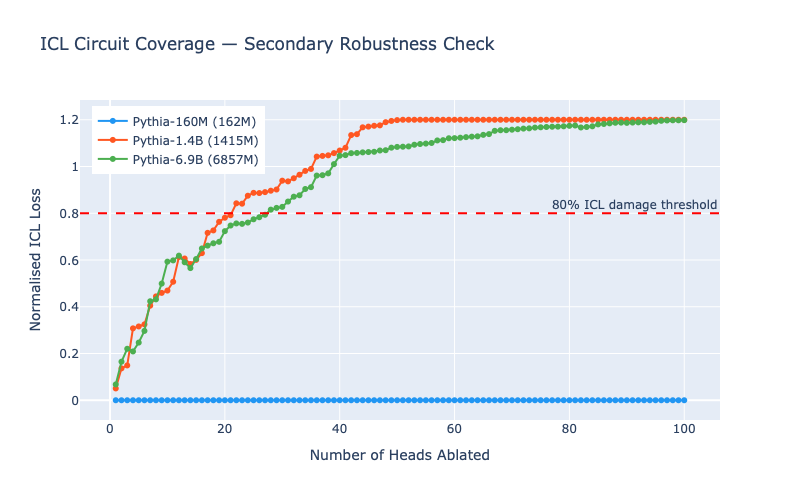}
\end{subfigure}
\hfill
\begin{subfigure}[t]{0.49\linewidth}
  \centering
  \includegraphics[width=\linewidth, trim=30 37 37 105, clip]{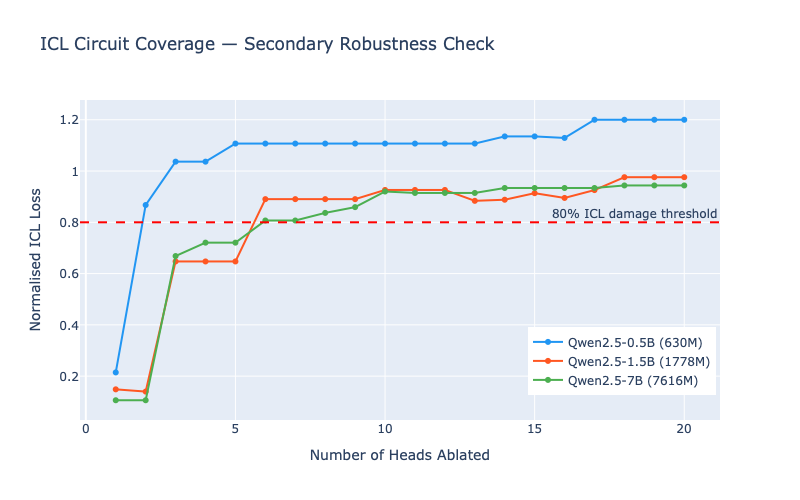}
\end{subfigure}
\caption{ICL ablation curves for Pythia (left) and Qwen2.5 (right).
Pythia-160M flatlines at zero, showing no functional induction heads.
Pythia-1.4B and 6.9B rise gradually and cross the threshold only after
many ablations. Qwen2.5 models cross 80\% within the first few
ablations and plateau, showing the circuit is carried by very few heads
regardless of scale.}
\label{fig:icl_curves}
\end{figure*}

Pythia-1.4B and 6.9B both solve the task but require 22 and 28
ablations respectively. The Pythia induction circuit becomes more
distributed as scale increases, not less. Within Qwen2.5, all three
models break in two to six ablations across a 14$\times$ parameter
range. Qwen2.5-7B at 7B parameters needs 6 ablations while Pythia-1.4B
at 1.4B needs 22. The concentration advantage of GQA holds even when
comparing a model five times larger against a smaller MHA baseline.
Figure~\ref{fig:icl_curves} shows the ablation curves for both
families.
 
\subsection{Factual Recall Undergoes a Phase Transition in GQA Models}
 
Table~\ref{tab:factual} shows factual recall results across both families and both analysis conditions.
 
\begin{table*}[t]
\centering
\caption{Factual recall results across Pythia and Qwen2.5 under
per-model and shared fact conditions. H-80\% is the number of heads
ablated to cause 80\% accuracy damage. Qwen2.5 circuit geometry is
identical across both conditions at every scale. Pythia circuit
geometry shifts substantially between per-model and shared facts,
reflecting sensitivity to fact difficulty.}
\label{tab:factual}
\small
\setlength{\tabcolsep}{4pt}
\begin{tabular}{lcccc}
\toprule
\textbf{Model} & \textbf{Facts (per)} & \textbf{Facts (shared)} & \textbf{Heads-to-80\% (per)} & \textbf{Heads-to-80\% (shared)} \\
\midrule
Pythia-160M  & 91  & 84  & 135 & 135 \\
Pythia-1.4B  & 265 & 84  & 36  & 50  \\
Pythia-6.9B  & 357 & 84  & 4   & 93  \\
\midrule
Qwen2.5-0.5B & 267 & 230 & 1   & 1   \\
Qwen2.5-1.5B & 299 & 230 & 1   & 1   \\
Qwen2.5-7B   & 330 & 230 & 1   & 1   \\
\bottomrule
\end{tabular}
\end{table*}
 
Pythia factual recall circuits are markedly diffuse. The shared
fact analysis requires 50 to 135 ablations to cause 80\% accuracy
damage. All three models show peak factual signal only in the final
layers \citep{meng2022locating}. Pythia-1.4B and Pythia-6.9B show
substantially different top heads and critical layers between per-model
and shared fact conditions. On shared (easier) facts, no single layer
causes more than 30\% damage. On harder per-model facts, the circuit
concentrates in mid-to-late layers and layer 16 causes 100\% damage
for Pythia-6.9B. This does not appear in Qwen2.5, which shows identical top
heads and layer profiles across both conditions.

\begin{figure*}[t]
\centering
\includegraphics[width=\linewidth,height=5cm,keepaspectratio, trim=5 5 5 17, clip]{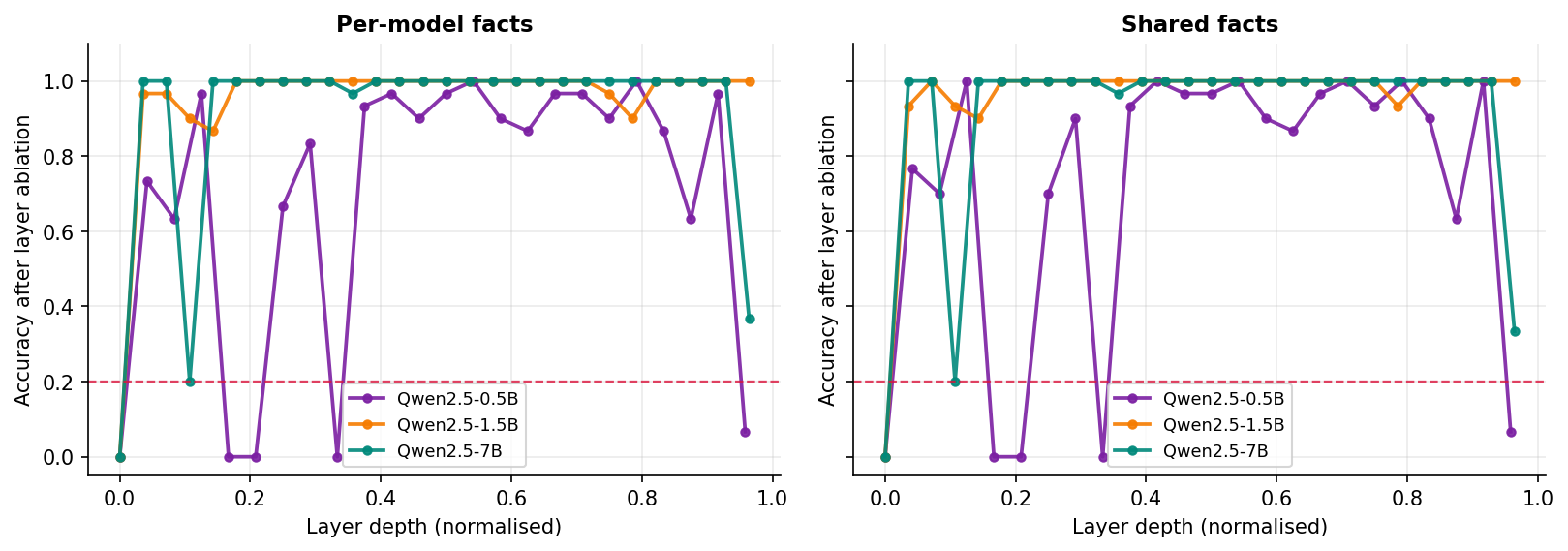}
\caption{Layer-wise ablation for Qwen2.5 factual recall. Each point shows accuracy after ablating all heads in that layer. Per-model facts (left) and shared facts (right) show identical profiles, showing the circuit location is stable across input conditions.}
\label{fig:qwen_layerwise}
\end{figure*}

All three Qwen2.5 models require exactly one ablation to cause 80\%
accuracy damage on both per-model and shared facts. A discrete phase
transition occurs between 0.5B and 1.5B. Qwen2.5-0.5B concentrates
factual recall at layer 4 and requires eight ablations to break.
Qwen2.5-1.5B and 7B concentrate entirely at layer 0 and break with a
single ablation. Figure~\ref{fig:qwen_layerwise} shows this directly: 
ablating all heads in layer 0 collapses accuracy to zero for Qwen2.5-1.5B 
and 7B while causing only moderate damage for Qwen2.5-0.5B. The profile is 
identical between per-model and shared fact conditions, showing that the 
Qwen2.5 circuit location is stable across fact difficulty.

Table~\ref{tab:factual_diagnostic} shows how the bottleneck breaks 
down at the KV head level: ablating the single top KV head at layer 0 
causes 97.2\% accuracy damage for Qwen2.5-1.5B and 92.8\% for Qwen2.5-7B, 
while the same ablation causes only 36.4\% damage for Qwen2.5-0.5B.

\begin{table}[H]
\centering
\caption{Layer 0 ablation diagnostic for Qwen2.5 factual recall. Acc. after L0 ablation is the remaining accuracy after ablating all attention heads in layer 0. Drop (\%) is the accuracy reduction when all attention heads at layer 0 are ablated simultaneously.}
\setlength{\tabcolsep}{4pt}
\label{tab:factual_diagnostic}
\small
\begin{tabular}{lccc}
\toprule
\textbf{Model} & \textbf{Top Head} & \textbf{Acc. after L0 Ablation} & \textbf{Drop (\%)} \\
\midrule
Qwen2.5-0.5B & L4 H7 & 0.345 & 36.4 \\
Qwen2.5-1.5B & L0 H6 & 0.017 & 97.2 \\
Qwen2.5-7B   & L0 H0 & 0.048 & 92.8 \\
\bottomrule
\end{tabular}
\end{table}
 
For Qwen2.5-1.5B, ablating KV head 1 at layer 0 alone collapses
accuracy to 0.017, while control ablations at any other layer improve
accuracy. Qwen2.5-7B shows the same structure. Qwen2.5-0.5B shows a
different structure: the top head is at layer 4 head 7 and ablating
both layer 0 and layer 4 together leaves 12.7\% residual accuracy,
pointing to a partial backup pathway. This dual bottleneck places 0.5B
below the phase transition threshold. On shared facts, no single layer
causes more than 30\% accuracy damage for Pythia-1.4B or Pythia-6.9B,
showing that MHA models have no comparable bottleneck structure.
 
Across all three circuit types, Qwen2.5 GQA models require one to six
ablations to cause 80\% circuit damage while Pythia MHA models require
4 to 135. At matched scales, Pythia-1.4B versus Qwen2.5-1.5B, the
difference is 50 versus 1 ablation for shared factual recall and 22
versus 6 for induction heads. Architecture drives this difference.
  
\section{Implications}
 
Architecture is a first-class variable for mechanistic
interpretability. GQA versus MHA predicts circuit concentration more
reliably than parameter count. Most deployed frontier models already
use GQA or similar KV-sharing mechanisms for inference efficiency.
Large deployed models may therefore be substantially more amenable to
circuit-level analysis than is commonly assumed. Interpretability tool
development should benchmark across architecture families rather than
model sizes alone. These results are encouraging for
interpretability-based safety monitoring. GQA circuits are stable
across fact difficulty while MHA circuits shift substantially between
input regimes. A monitoring tool built on a GQA circuit will behave
consistently across inputs while a tool built on an MHA circuit may
not. Identifying circuits is necessary but not sufficient for reliable
oversight. Circuits must also be verified to remain stable under
deployment conditions.

\section{Limitations}
 
Our fact set covers world geography, science, history, and culture.
The factual recall findings may not generalise to facts requiring
multi-hop reasoning or temporal context. We cannot fully separate
architecture from training. Pythia and Qwen2.5 differ in training
data, tokenizer, and training recipe in addition to attention
mechanism. The GQA hypothesis is the most parsimonious explanation but
a controlled experiment with matched models trained with and without
GQA would be needed to isolate the architectural effect. Our study
covers three circuit types chosen for their prior literature support.
Whether the architecture-driven concentration pattern holds for
circuits underlying safety-relevant behaviours such as deception or
goal-directed reasoning remains an open question.
 
\section{Conclusion}
 
Mechanistic interpretability difficulty is not a monotone function of
model size. Across three circuit types and six models, we find that
the attention mechanism determines circuit geometry more reliably than
parameter count. GQA models produce circuits that concentrate into one
or two heads, remain stable across input conditions, and break cleanly
under targeted ablation. MHA models at comparable scales produce
circuits that are diffuse, input-sensitive, and resistant to surgical
intervention. GQA was designed for inference efficiency. Its effect on
circuit tractability is a consequence of structural constraints on
value-space computation, not an intended property. This architectural
choice incidentally produces more interpretable models at scale. The
field should take this into account when deciding which models to
prioritise for study and which design choices to encourage.
 
\section*{Acknowledgements}
We thank the mechanistic interpretability community for open-sourcing TransformerLens. 
Experiments were conducted using publicly available models from Hugging Face. 
We also thank RunPod for providing compute resources.
 
\section*{Impact Statement}
This work studies how attention architecture affects mechanistic
interpretability. GQA, already widely adopted for inference efficiency,
incidentally produces more tractable circuits at scale. We hope this
encourages interpretability research on deployed models and evaluation
of tools across architecture families. More broadly, a better
understanding of what makes models interpretable at the architectural
level may inform safer model design choices and support the development
of reliable oversight tools for deployed AI systems. All model weights,
datasets and libraries used in this work are publicly available and
were used in accordance with their respective licenses and terms of use.

\bibliography{example_paper}
\bibliographystyle{icml2026}

\newpage
\appendix
\onecolumn

\section{Factual Recall Dataset: Domain Coverage and Prompt Format}

\begin{table}[H]
\centering
\caption{Five sample prompts per domain from the 493-fact set. Answers are single tokens.}
\label{tab:app_facts}
\small
\setlength{\tabcolsep}{4pt}
\begin{tabular}{>{\columncolor[gray]{0.9}}p{2.8cm} p{8.5cm}}
\toprule
\textbf{Domain} & \textbf{Sample Prompts (answer)} \\
\midrule
World Geography &
The capital of France is (\textit{Paris}) \\
& The Nile River is located in (\textit{Egypt}) \\
& The Eiffel Tower is located in (\textit{Paris}) \\
& The largest ocean in the world is the (\textit{Pacific}) \\
& The largest country by area is (\textit{Russia}) \\
\midrule
Science \& Chemistry &
The chemical symbol for gold is (\textit{Au}) \\
& The largest organ in the human body is the (\textit{skin}) \\
& The atomic number of hydrogen is (\textit{1}) \\
& The SI unit of length is (\textit{meter}) \\
& The hardest natural material is (\textit{diamond}) \\
\midrule
History &
World War II ended in (\textit{1945}) \\
& The French Revolution began in (\textit{1789}) \\
& The Berlin Wall fell in (\textit{1989}) \\
& The first moon landing occurred in (\textit{1969}) \\
& India gained independence in (\textit{1947}) \\
\midrule
Notable People &
Marie Curie was born in (\textit{Poland}) \\
& Isaac Newton was born in (\textit{England}) \\
& Aristotle was a student of (\textit{Plato}) \\
& Albert Einstein was born in (\textit{Germany}) \\
& Mahatma Gandhi was born in (\textit{India}) \\
\midrule
Literature \& Arts &
The author of 1984 is (\textit{George}) \\
& The Mona Lisa was painted by (\textit{Leonardo}) \\
& The Starry Night was painted by (\textit{Vincent}) \\
& Swan Lake was composed by (\textit{Tchaikovsky}) \\
& The author of Hamlet is (\textit{William}) \\
\midrule
Technology &
Apple was founded by (\textit{Steve}) \\
& Linux was created by (\textit{Linus}) \\
& The iPhone was first released in (\textit{2007}) \\
& Python was created by (\textit{Guido}) \\
& Google was founded by (\textit{Larry}) \\
\midrule
Sports &
The FIFA World Cup is held every (\textit{four}) years \\
& Basketball was invented by (\textit{James}) \\
& Golf originated in (\textit{Scotland}) \\
& Cricket originated in (\textit{England}) \\
& Wimbledon is held in (\textit{London}) \\
\midrule
Food \& Culture &
Pizza originated in (\textit{Italy}) \\
& Sushi originated in (\textit{Japan}) \\
& Coffee originated in (\textit{Ethiopia}) \\
& Vodka originated in (\textit{Russia}) \\
& Champagne is produced in (\textit{France}) \\
\midrule
Mythology \& Religion &
The king of Greek gods is (\textit{Zeus}) \\
& The Norse god of thunder is (\textit{Thor}) \\
& The sacred text of Islam is the (\textit{Quran}) \\
& Islam was founded by (\textit{Muhammad}) \\
& The holy city of Islam is (\textit{Mecca}) \\
\midrule
Currencies \& Languages &
The currency of Japan is the (\textit{yen}) \\
& The currency of India is the (\textit{rupee}) \\
& The official language of China is (\textit{Mandarin}) \\
& The official language of Brazil is (\textit{Portuguese}) \\
& The most spoken language is (\textit{Mandarin}) \\
\bottomrule
\end{tabular}
\end{table}
Table~\ref{tab:app_facts} shows five representative prompts from each domain 
in the 493-fact set. All prompts follow a natural cloze completion format 
and require single-token answers. The fact set spans ten domains chosen to 
ensure diverse coverage of world knowledge while avoiding multi-hop reasoning or temporal context.

\section{Additional Results}
\label{app:results}
 
\subsection{Top Contributing IOI Heads per Model}

Table~\ref{tab:app_top_heads} shows the top-5 IOI contributing
heads for each model. For Qwen2.5-1.5B all five top heads are at layer
0 and for Qwen2.5-7B all five are also at layer 0. The GQA row
structure is visible directly: because KV heads are shared across query
heads, ablating one KV head ablates an entire row of query heads
simultaneously, concentrating the damage at the shared layer. For
Pythia, the top-5 heads are distributed across multiple layers with no
dominant layer. Across all six models, the top-scoring IOI head and the
top-scoring ICL head are different individual heads, yet the circuit
concentration pattern is consistent within each architecture family
across both tasks. The architecture-driven concentration effect is a
property of how the architecture organises computation, not of any
single head.
 
\begin{table}[h]
\centering
\caption{Top-5 IOI contributing heads per model by (layer, head) index.}
\label{tab:app_top_heads}
\small
\begin{tabular}{ll}
\toprule
\textbf{Model} & \textbf{Top-5 heads (layer, head)} \\
\midrule
Pythia-160M  & (8,9), (7,11), (3,3), (1,6), (11,3) \\
Pythia-1.4B  & (12,8), (10,7), (0,2), (23,13), (15,7) \\
Pythia-6.9B  & (16,30), (24,26), (27,30), (25,24), (23,18) \\
\midrule
Qwen2.5-0.5B & (23,0), (23,1), (23,2), (23,3), (23,4) \\
Qwen2.5-1.5B & (0,6), (0,7), (0,8), (0,9), (0,10) \\
Qwen2.5-7B   & (0,1), (0,2), (0,3), (0,4), (0,5) \\
\bottomrule
\end{tabular}
\end{table}

\subsection{Pythia Fact-Difficulty Diagnostic}
 
Table~\ref{tab:app_pythia_diag} reports the top-scoring head and mean logit
gap for Pythia-6.9B and Pythia-1.4B under per-model and shared fact
conditions. The top head changes substantially between conditions for both
models and the logit gap is higher on shared (easier) facts for both. This
supports the fact-difficulty dependent circuit geometry claim in the main
results.
 
\begin{table}[H]
\centering
\caption{Pythia diagnostic results comparing per-model and shared fact
conditions. Top head layer shows where the primary contributing head is
located under each condition.}
\label{tab:app_pythia_diag}
\small
\begin{tabular}{llcccc}
\toprule
\textbf{Model} & \textbf{Condition} & \textbf{Logit Gap} &
\textbf{Top Head Layer} & \textbf{Top Head Score} \\
\midrule
\multirow{2}{*}{Pythia-1.4B}
  & Per-model & 1.006 & L12 & 0.903 \\
  & Shared    & 1.202 & L23 & 0.552 \\
\midrule
\multirow{2}{*}{Pythia-6.9B}
  & Per-model & 0.647 & L4  & 0.655 \\
  & Shared    & 1.098 & L16 & 0.827 \\
\bottomrule
\end{tabular}
\end{table}

\subsection{IOI Head Contribution Score Heatmaps}
 
\begin{figure}[h]
\centering
\begin{subfigure}[t]{0.32\linewidth}
\centering
\includegraphics[width=\linewidth, trim=37 37 23 97, clip]{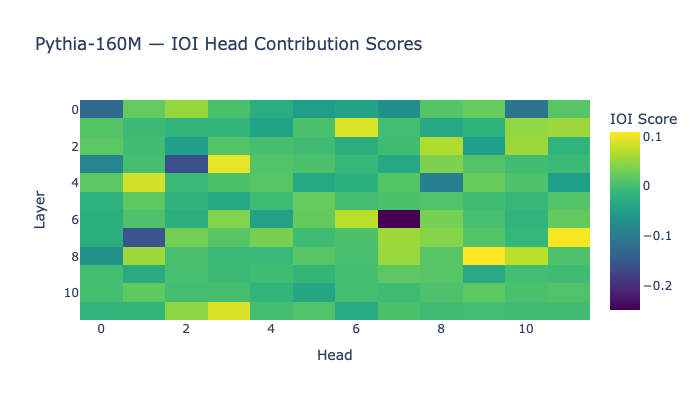}
\subcaption{\textcolor[HTML]{1565C0}{\textbf{Pythia-160M}}}
\end{subfigure}
\hfill
\begin{subfigure}[t]{0.32\linewidth}
\centering
\includegraphics[width=\linewidth, trim=37 37 23 97, clip]{figures/pythia-ioi.png}
\subcaption{\textcolor[HTML]{1565C0}{\textbf{Pythia-1.4B}}}
\end{subfigure}
\hfill
\begin{subfigure}[t]{0.32\linewidth}
\centering
\includegraphics[width=\linewidth, trim=37 37 23 97, clip]{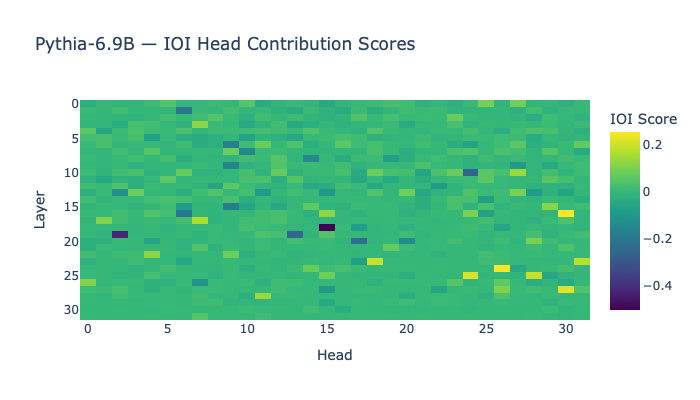}
\subcaption{\textcolor[HTML]{1565C0}{\textbf{Pythia-6.9B}}}
\end{subfigure}
 
\vspace{0.3cm}
 
\begin{subfigure}[t]{0.32\linewidth}
\centering
\includegraphics[width=\linewidth, trim=37 37 23 97, clip]{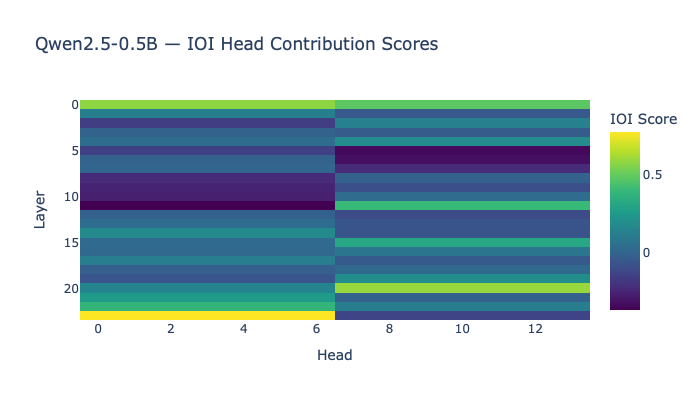}
\subcaption{\textcolor[HTML]{B71C1C}{\textbf{Qwen2.5-0.5B}}}
\end{subfigure}
\hfill
\begin{subfigure}[t]{0.32\linewidth}
\centering
\includegraphics[width=\linewidth, trim=37 37 23 97, clip]{figures/qwen-ioi.png}
\subcaption{\textcolor[HTML]{B71C1C}{\textbf{Qwen2.5-1.5B}}}
\end{subfigure}
\hfill
\begin{subfigure}[t]{0.32\linewidth}
\centering
\includegraphics[width=\linewidth, trim=37 37 23 97, clip]{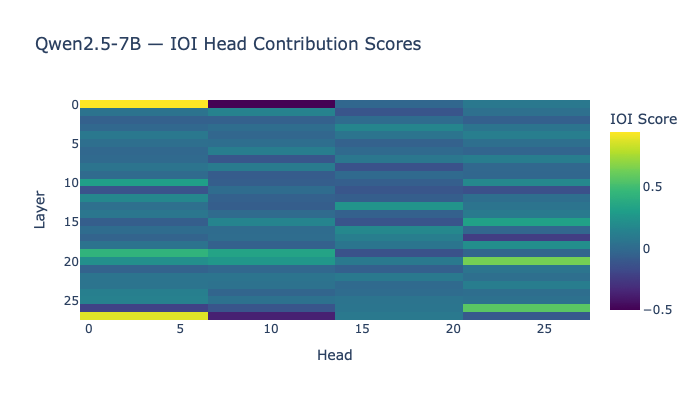}
\subcaption{\textcolor[HTML]{B71C1C}{\textbf{Qwen2.5-7B}}}
\end{subfigure}
 
\caption{IOI head contribution score heatmaps across all six models. Each
cell shows the logit diff drop when that (layer, head) pair is ablated.
\textcolor[HTML]{1565C0}{\textbf{Pythia (MHA)}} shows contributions
scattered across many layers and heads with no dominant structure.
\textcolor[HTML]{B71C1C}{\textbf{Qwen2.5 (GQA)}} shows a single
bright band at layer 0 for 1.5B and 7B and at layer 23 for 0.5B, reflecting
the phase transition between these scales.}
\label{fig:app_ioi_heatmaps}
\end{figure}

Figure~\ref{fig:app_ioi_heatmaps} shows IOI head contribution score heatmaps
for all six models. The architecture contrast is visible at every scale:
Pythia heatmaps show diffuse scattered contributions while Qwen2.5 heatmaps
show a single concentrated band.
 
\subsection{ICL Induction Head Score Heatmaps}
 
\begin{figure}[h]
\centering
\begin{subfigure}[t]{0.32\linewidth}
\centering
\includegraphics[width=\linewidth, trim=37 37 23 97, clip]{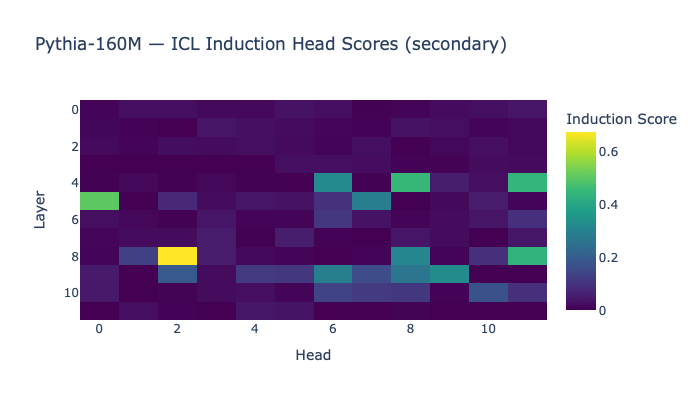}
\subcaption{\textcolor[HTML]{1565C0}{\textbf{Pythia-160M}}}
\end{subfigure}
\hfill
\begin{subfigure}[t]{0.32\linewidth}
\centering
\includegraphics[width=\linewidth, trim=37 37 23 97, clip]{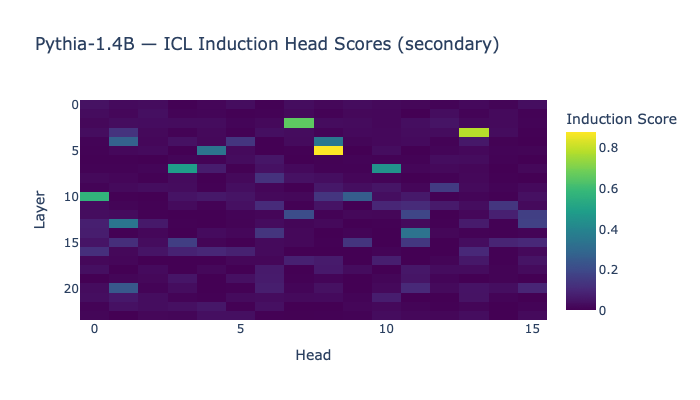}
\subcaption{\textcolor[HTML]{1565C0}{\textbf{Pythia-1.4B}}}
\end{subfigure}
\hfill
\begin{subfigure}[t]{0.32\linewidth}
\centering
\includegraphics[width=\linewidth, trim=37 37 23 97, clip]{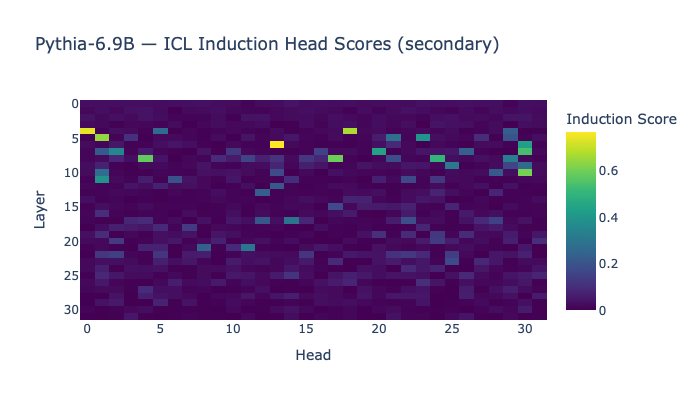}
\subcaption{\textcolor[HTML]{1565C0}{\textbf{Pythia-6.9B}}}
\end{subfigure}
 
\vspace{0.3cm}
 
\begin{subfigure}[t]{0.32\linewidth}
\centering
\includegraphics[width=\linewidth, trim=37 37 23 97, clip]{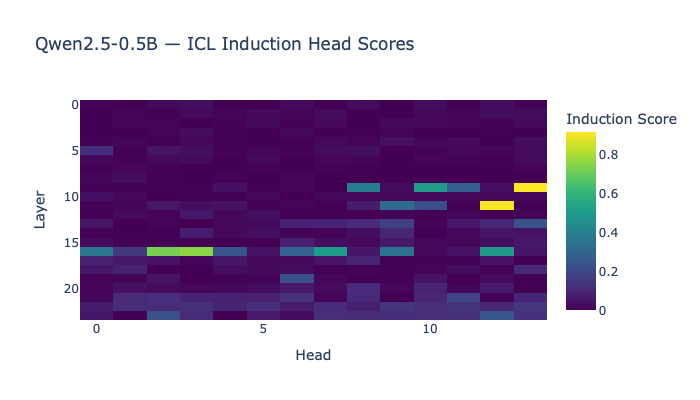}
\subcaption{\textcolor[HTML]{B71C1C}{\textbf{Qwen2.5-0.5B}}}
\end{subfigure}
\hfill
\begin{subfigure}[t]{0.32\linewidth}
\centering
\includegraphics[width=\linewidth, trim=37 37 23 97, clip]{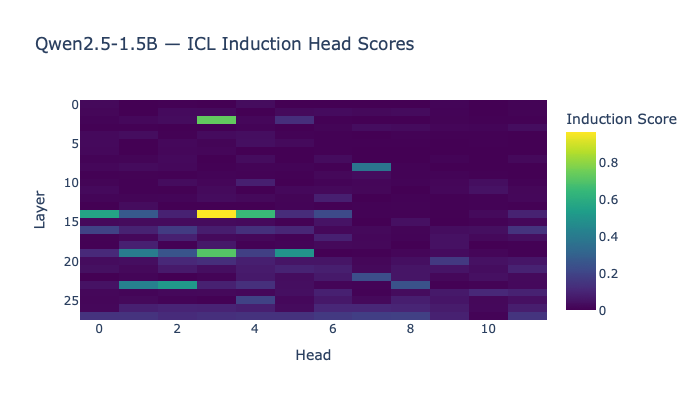}
\subcaption{\textcolor[HTML]{B71C1C}{\textbf{Qwen2.5-1.5B}}}
\end{subfigure}
\hfill
\begin{subfigure}[t]{0.32\linewidth}
\centering
\includegraphics[width=\linewidth, trim=37 37 23 97, clip]{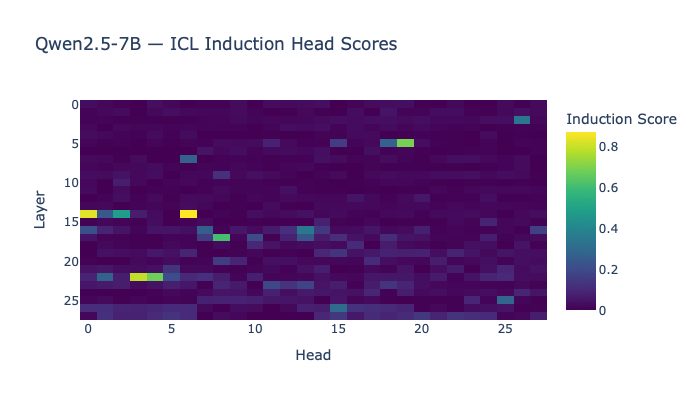}
\subcaption{\textcolor[HTML]{B71C1C}{\textbf{Qwen2.5-7B}}}
\end{subfigure}
 
\caption{ICL induction head score heatmaps across all six models on
random repeated-token sequences. Each cell shows the mean attention
weight at the induction offset position for that (layer, head) pair.
\textcolor[HTML]{1565C0}{\textbf{Pythia (MHA)}} shows
induction scores scattered across the full layer-head matrix.
Pythia-160M has one dominant cell at layer 8 but Pythia-1.4B and 6.9B
show increasing scatter with no clearly dominant layer and the number
of high-scoring heads grows with scale.
\textcolor[HTML]{B71C1C}{\textbf{Qwen2.5 (GQA)}} shows
induction scores concentrated in specific mid-to-late layer bands.
Qwen2.5-1.5B shows a bright cluster around layers 14-20 with fewer
active heads. Qwen2.5-7B shows a similar mid-network concentration,
with notably fewer high-scoring cells than Pythia-6.9B at comparable
scale.}
\label{fig:app_icl_heatmaps}
\end{figure}

Figure~\ref{fig:app_icl_heatmaps} shows ICL induction head score heatmaps
for all six models on the secondary random repeated-token task. The same
architecture contrast holds: Pythia heatmaps show scattered induction scores
across the full layer-head matrix while Qwen2.5 heatmaps show a smaller
number of dominant heads.
 
\end{document}